\documentclass[11pt]{article}

\usepackage[final]{acl}

\usepackage{times}
\usepackage{latexsym}

\usepackage[T1]{fontenc}

\usepackage[utf8]{inputenc}

\usepackage{microtype}

\usepackage{inconsolata}

\usepackage{graphicx}

\usepackage{amsmath}
\usepackage{amssymb}

\usepackage{booktabs}
\usepackage{multirow}

\usepackage{xspace}
\DeclareRobustCommand{\ppo}{\texttt{PPO}\xspace}
\DeclareRobustCommand{\grpo}{\texttt{GRPO}\xspace}
\DeclareRobustCommand{\dapo}{\texttt{DAPO}\xspace}
\DeclareRobustCommand{\depo}{\texttt{DEPO}\xspace}
\DeclareRobustCommand{\polaris}{\texttt{Polaris}\xspace}
\DeclareRobustCommand{\eg}{e.g.,\xspace}
\DeclareRobustCommand{\ie}{i.e.,\xspace}
\title{Difficulty-Estimated Policy Optimization}

\author{
  \textbf{Yu Zhao\textsuperscript{1}\thanks{Equal contribution.}},
  \textbf{Fan Jiang\textsuperscript{3}\footnotemark[1]},
  \textbf{Liu TianLe\textsuperscript{2}\footnotemark[1]},
  \textbf{Bo Zeng\textsuperscript{1}},
\\
  \textbf{Xintong Wang\textsuperscript{1}},
  \textbf{Jiaxuan Teng\textsuperscript{4}},
  \textbf{Yu Liu\textsuperscript{2}\thanks{Corresponding authors.}},
  \textbf{Yichao Du\textsuperscript{4}\footnotemark[2]}
\\
\\
  \textsuperscript{1}Alibaba Group
\\
  \textsuperscript{2}School of Software Technology, Dalian University of Technology, Dalian, China
\\
  \textsuperscript{3}Amazon
\\
  \textsuperscript{4}Independent Researcher
\\
  \small{
    \texttt{zhaoyu@sniper97.cn},
    \texttt{yuliu@dlut.edu.cn}
  }
}

\begin{document}

\maketitle

\begin{abstract}
Recent advancements in Large Reasoning Models (LRMs), exemplified by DeepSeek-R1, have underscored the potential of scaling inference-time compute through Group Relative Policy Optimization (\grpo). However, \grpo frequently suffers from gradient signal attenuation when encountering problems that are either too trivial or overly complex. In these scenarios, the disappearance of inter-group advantages makes the gradient signal susceptible to noise, thereby jeopardizing convergence stability. While variants like \dapo attempt to rectify gradient vanishing, they do not alleviate the substantial computational overhead incurred by exhaustive rollouts on low-utility samples. In this paper, we propose \textbf{D}ifficulty-\textbf{E}stimated \textbf{P}olicy \textbf{O}ptimization (\textbf{\depo}), a novel framework designed to optimize the efficiency and robustness of reasoning alignment. \depo integrates an online Difficulty Estimator that dynamically assesses and filters training data before the rollout phase. This mechanism ensures that computational resources are prioritized for samples with high learning potential. Empirical results show that DEPO achieves approximately 1.86× faster rollouts than \dapo in the profiled configuration, while improving benchmark performance over \grpo at comparable training cost. Our approach significantly lowers the computational barrier for training high-performance reasoning models, offering a more sustainable path for reasoning scaling.
\end{abstract}

\section{Introduction}
The recent emergence of Large Reasoning Models (LRMs), exemplified by OpenAI's o1 series~\citep{openai2024openaio1card, Guo_2025, yin-etal-2025-marco}, represents a transformative shift toward models capable of executing complex, multi-step cognitive chains. This progress is largely driven by the adoption of Reinforcement Learning from Verifiable Rewards (RLVR). In contrast to conventional Reinforcement Learning from Human Feedback (RLHF), which is often constrained by the subjectivity and variability of human preferences, RLVR leverages objective supervisory signals, such as mathematical correctness or code execution outcomes, to provide deterministic, automated feedback~\citep{lambert2025tulu, lightman2024lets}.

\begin{figure}[t]
    \centering
    \setlength{\abovecaptionskip}{-0.05cm}
    \setlength{\belowcaptionskip}{-0.45cm}
    \includegraphics[width=0.85\columnwidth]{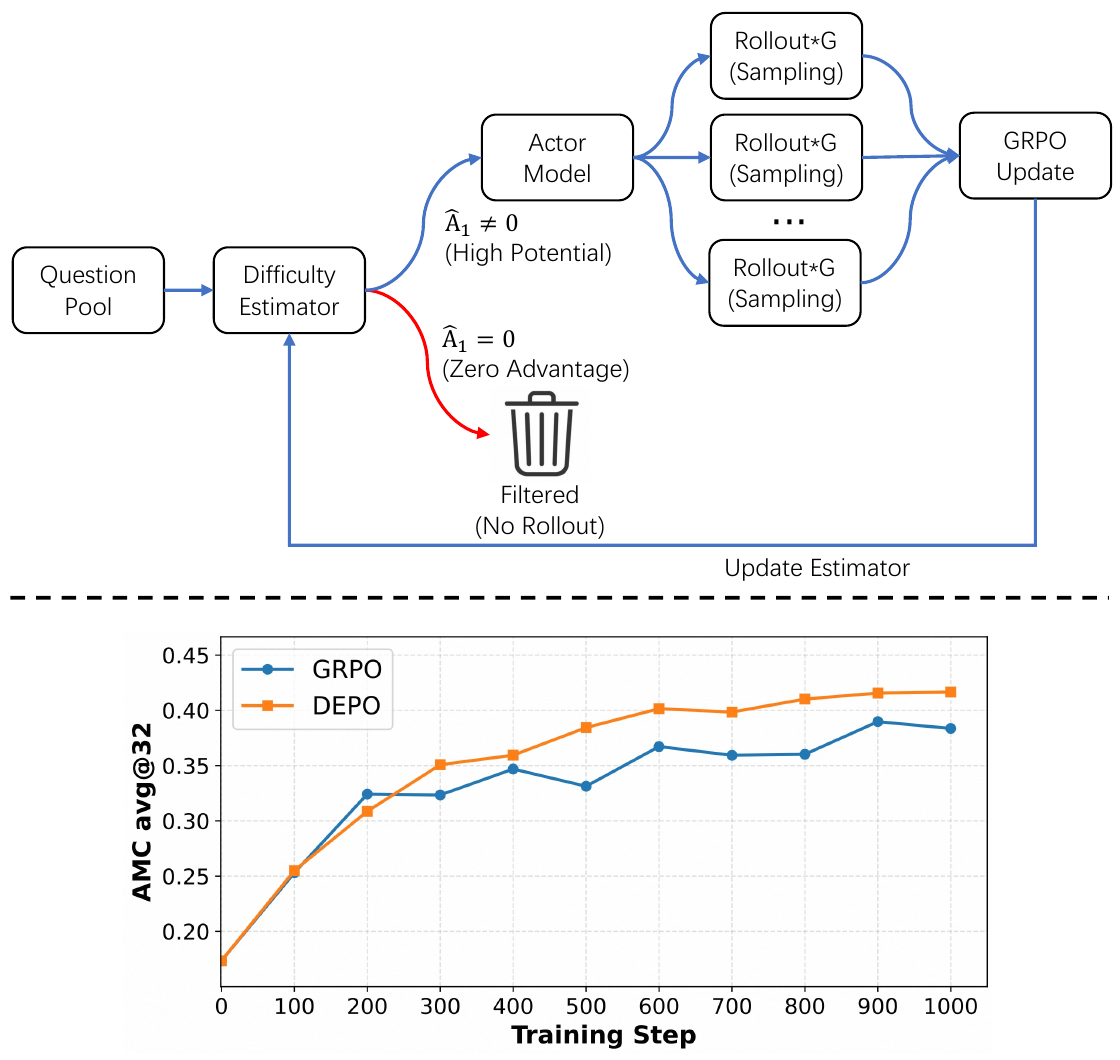}
    \caption{Top: the overview of our proposed \depo framework. Bottom: Training dynamics of downstream accuracy of \grpo and \depo.}
    \label{fig:main_fig_and_grpo_vs_depo_acc}
\end{figure}

Among the algorithms driving RLVR, Group Relative Policy Optimization (\grpo)~\citep{shao2024deepseekmathpushinglimitsmathematical} has emerged as a robust alternative to traditional Proximal Policy Optimization (\ppo)~\citep{schulman2017proximalpolicyoptimizationalgorithms}. By eliminating the requirement for a value model and instead computing advantages relative to a group mean, \grpo significantly mitigates training instability. However, this stability imposes a prohibitive computational burden, as it necessitates the generation of multiple responses for every input. While existing research has explored efficiency via decoding-side optimizations~\citep{anonymous2026a, ma-etal-2025-cot, C3oT} or framework-level refinements (\eg REINFORCE~\citep{ahmadian-etal-2024-back}, \dapo~\citep{yu2025dapoopensourcellmreinforcement}, and PREPO~\citep{sun2025efficientreinforcementlearninglarge}), a critical bottleneck persists: the rollout inefficiency stemming from an imbalance in sample difficulty.

During \grpo training, a disproportionate portion of the computational budget is spent on ``rollouts'' (sampling responses). However, the utility of these rollouts is often undermined by samples at the extremes of the difficulty spectrum: those that are either trivial or intractable yield negligible advantage signals. This leads to vanishing gradients and significant computational waste~\citep{yu2025dapoopensourcellmreinforcement, sun2025efficientreinforcementlearninglarge}. While existing mitigation strategies, such as dynamic sampling~\citep{yu2025dapoopensourcellmreinforcement}, PPL-based re-ranking~\citep{sun2025efficientreinforcementlearninglarge}, or offline filtering~\citep{Polaris2025}, attempt to address this bottleneck, they remain suboptimal. Specifically, offline methods overlook the fact that a sample's difficulty is a moving target that shifts as the actor model evolves. Meanwhile, online re-ranking methods often introduce significant latency or exhibit sensitivity to stochastic noise.

To address these challenges, we propose \textbf{\depo} (\textbf{D}ifficulty \textbf{E}stimated \textbf{P}olicy \textbf{O}ptimization), an efficient online prompt filtering algorithm.
As shown in Figure~\ref{fig:main_fig_and_grpo_vs_depo_acc}, we introduce a lightweight Difficulty Estimator integrated seamlessly into the \grpo pipeline. Our approach employs a BERT-based encoder~\citep{devlin-etal-2019-bert} equipped with two specialized prediction heads to estimate a prompt's difficulty in real-time to facilitate dynamic data filtering. Notably, the estimator is updated synchronously along with the actor model by leveraging the trajectories (\ie rewards and log-probabilities) generated during the standard \grpo training loop. This design avoids separately collecting difficulty-supervision data and running additional reference-set rollouts for candidate scoring. The estimator is updated using feedback from the ongoing policy-training loop to adapt to the actor’s evolving capabilities.

By prioritizing prompts with intermediate predicted success probabilities, DEPO avoids rollout generation for candidates outside the selected difficulty range.Figure~\ref{fig:main_fig_and_grpo_vs_depo_acc} shows improved downstream AMC23 Avg@32 performance over GRPO at the evaluated checkpoints . Empirical evaluations demonstrate that \depo provides the following advantages:
\begin{itemize}
    \item \textbf{Dynamic Online Filtering}: \depo filters training instances in an online fashion, capturing the temporal dynamics of sample difficulty relative to the actor's evolving policy.
    \item \textbf{Superior Performance-Efficiency Trade-off}: Our approach outperforms \grpo by 1.5\% across multiple mathematical reasoning benchmarks while maintaining comparable training efficiency.
    \item \textbf{Framework Orthogonality}: As a plug-and-play optimization, \depo is orthogonal to existing state-of-the-art frameworks such as \dapo. When integrated, it achieves up to a 2.4\% accuracy improvement while simultaneously yielding a 41.4\% reduction in total computational overhead.
\end{itemize}

\section{Preliminary}

\subsection{Group Relative Policy Optimization (GRPO)}
Proximal Policy Optimization (\ppo)~\citep{schulman2017proximalpolicyoptimizationalgorithms} stabilizes policy gradient training with a clipped surrogate objective that bounds the probability ratio between the new policy $\pi_{\theta}$ and the old policy $\pi_{\theta_{\text{old}}}$, and estimates the advantage $A_t$ with a separate value model via Generalized Advantage Estimation~\citep{schulman2018highdimensionalcontinuouscontrolusing}. \grpo~\citep{shao2024deepseekmathpushinglimitsmathematical} removes that value model, replacing it with a \emph{group-based relative estimation}. For each input query $q$, the algorithm first samples a group of responses $\{o_i\}_{i=1}^G$ from the current policy $\pi_{\theta_{\text{old}}}$ and calculate the rewards for all outputs in the group $\{R_i\}_{i=1}^G$. Finally, it computes the advantage of each output by comparing its reward against the average reward of the entire group:
\begin{equation*}
    A_{i,t} = \frac{r_i - \operatorname{mean}(\{R_i\}_{i=1}^G)}{\operatorname{std}(\{R_i\}_{i=1}^G)}
\end{equation*}
\grpo also adopts the clipped surrogate objective, together with an additional KL penalty term to prevent the model from diverging too far from a reference policy:
\begin{align*}
    \mathcal{J}_{\grpo}(\theta) = {} & \mathbb{E}_{(q, a) \sim \mathcal{D},\, \{o_i\}_{i=1}^G \sim \pi_{\theta_{\text{old}}}(\cdot \mid q)} \\
    & \frac{1}{G} \sum_{i=1}^G \Big[ \min \Big( r_i(\theta) A_i, \\
    & \quad \operatorname{clip}\big( r_i(\theta), 1-\epsilon, 1+\epsilon \big) A_i \Big) \\
    & \quad - \beta \mathbb{D}_{\text{KL}}(\pi_{\theta} \,\|\, \pi_{\text{ref}}) \Big]
\end{align*}
Dropping the value model saves compute and memory, but leaves \grpo susceptible to gradient sparsity whenever intra-group rewards lack variance: if every sampled response for a prompt earns an identical reward, the within-group advantages vanish and optimization stalls on that prompt. This ``zero-variance'' problem worsens over training, as the share of prompts yielding uniform maximum reward grows with policy convergence~\citep{yu2025dapoopensourcellmreinforcement}, shrinking the effective batch size and amplifying gradient variance.

\begin{figure*}[t]
    \centering
    \setlength{\belowcaptionskip}{-0.3cm}
    \includegraphics[width=0.75\textwidth]{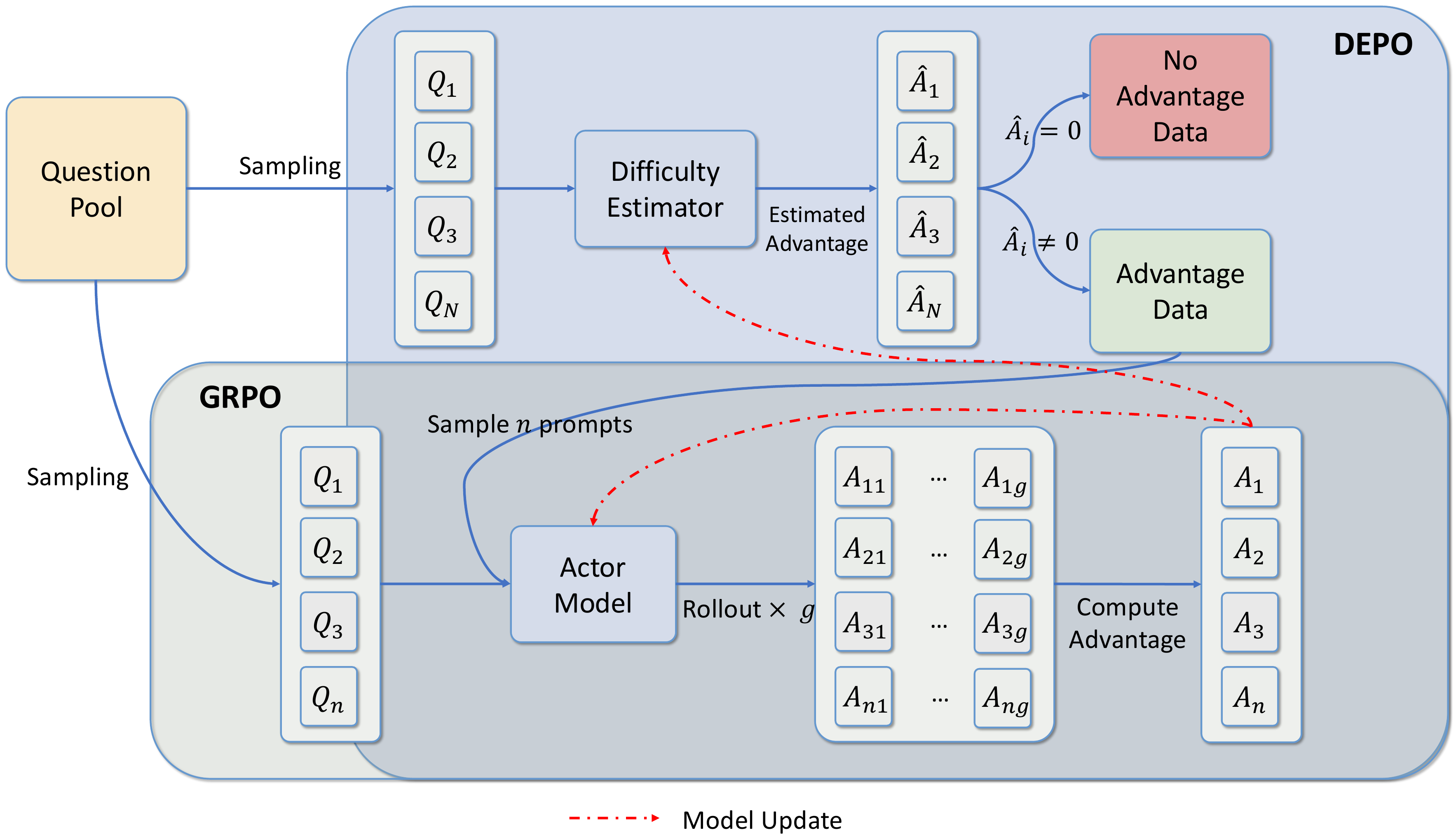}
    \caption{Architectural overview of the proposed \depo algorithm. \depo utilizes a Difficulty Estimator to predict an advantage-related filtering score $\hat{A}_i$ for each sampled question. In implementation, $\sigma(\hat{A}_i)$ estimates the prompt-level Avg@$G$ score, and prompts whose predicted Avg@$G$ is close to $0$ or $1$ are filtered before rollout. Samples with intermediate predicted scores are employed for updating the Actor Model using the standard \grpo algorithm. The Difficulty Estimator is simultaneously updated using the Avg@$G$ targets obtained from the \grpo rollouts.}
    \label{fig:depo}
\end{figure*}

\subsection{Existing Methods for Mitigating the Zero-Variance Problem of \grpo}
One mitigation is Dynamic Sampling via oversampling~\citep{yu2025dapoopensourcellmreinforcement}: each iteration oversamples prompts, evaluates rewards, discards those with uniform rewards, and optimizes only on the rest. This does remove zero-variance noise and stabilize training, but at a steep cost: full, high-latency rollouts must still be run for the enlarged sample set, inflating per-step computational overhead.

Alternatively, \citet{Polaris2025} curates data offline, periodically reconfiguring the training distribution in discrete stages by pruning trivial and intractable prompts; at each stage the current policy runs full rollouts over the training set to retain the challenging-but-solvable samples. This sustains a learning signal, but because the policy updates continuously while the data distribution refreshes only at sparse intervals, the \emph{offline} estimates lag the model's evolving capability.

\begin{figure}[t]
    \centering
    \setlength{\belowcaptionskip}{-0.3cm}
    \includegraphics[width=0.6\columnwidth]{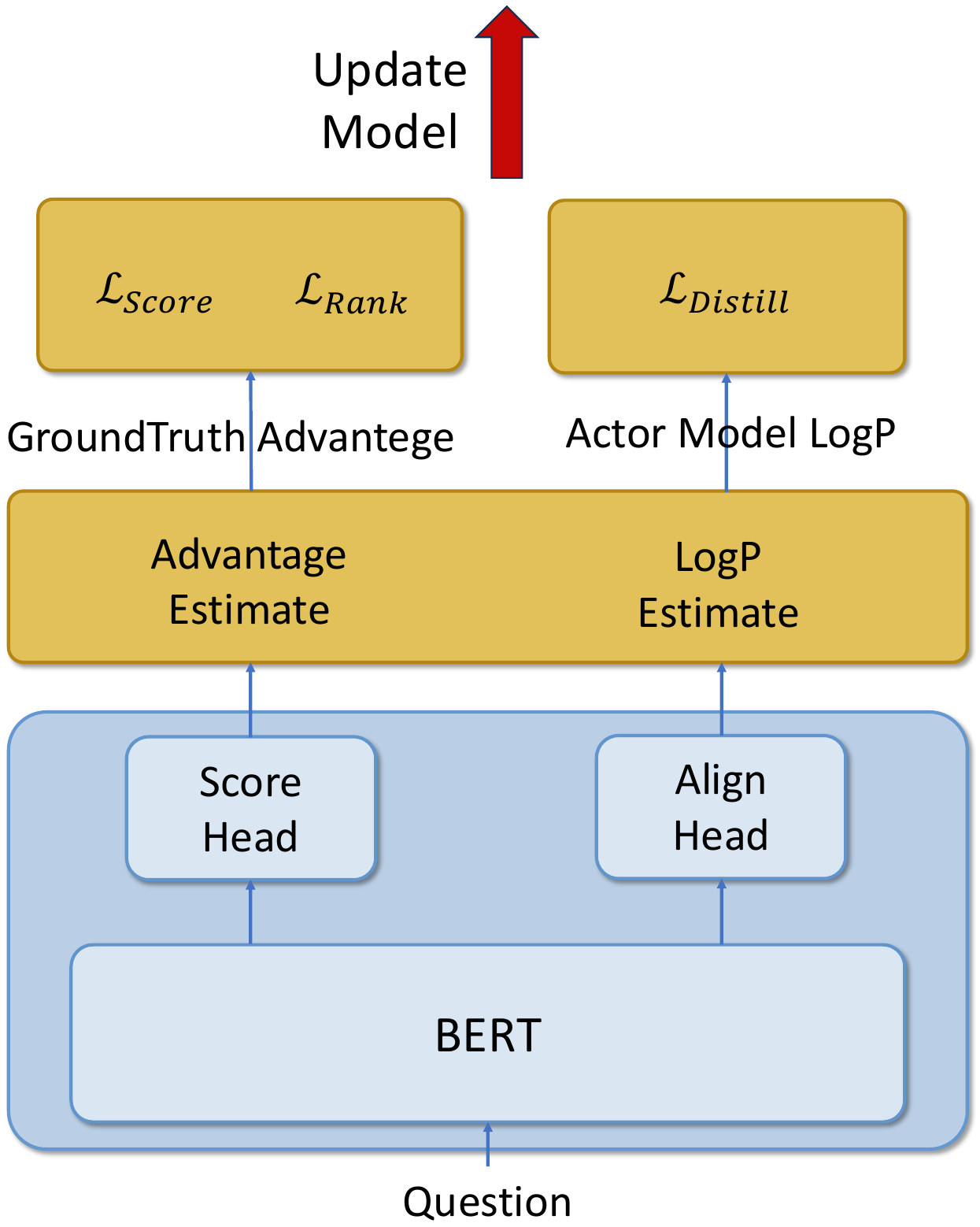}
    \caption{The online difficulty estimator architecture.}
    \label{fig:estimator}
\end{figure}

\section{\depo}

A key inefficiency in \grpo arises when all responses sampled for a prompt receive identical rewards, yielding no group-relative reward learning signal. Post-rollout filtering, such as dynamic sampling in \dapo, identifies these groups only after paying their generation cost. depo (\textbf{D}ifficulty \textbf{E}stimated \textbf{P}olicy \textbf{O}ptimization) instead scores candidate prompts with an online difficulty estimator before rollout and retains those with intermediate predicted success probabilities for actor training, as illustrated in Figure~\ref{fig:depo}.

\subsection{Online Difficulty Estimator}\label{sec:estimator}
Our Difficulty Estimator predicts the difficulty of a question \emph{from the perspective of the current actor}. Unlike static curriculum learning or offline filtering, it is trained continuously alongside the actor, so its estimates track the actor's evolving capability throughout RL training.

\subsubsection{Model Architecture}
As shown in Figure~\ref{fig:estimator}, the Difficulty Estimator is built upon a pre-trained BERT model~\citep{devlin-etal-2019-bert}. For any given question, the model takes the raw text of the question as input and is tasked with predicting two target values:
\begin{enumerate}
    \item \textbf{Estimated advantage-related filtering score}: A normalized score $\hat{A}\in[0,1]$ representing the prompt-level advantage potential of a given question. Rather than predicting the per-response GRPO advantage $A_{i,t}$ directly, we use the empirical Avg@$G$ reward as a tractable supervision signal. Predicted success probabilities near 0 or 1 indicate a greater likelihood of all-incorrect or all-correct response groups, respectively, motivating the selection of prompts with intermediate predicted probabilities.
    \item \textbf{Actor Perplexity (PPL)}: The perplexity of the current actor model when modeling the question contexts.
\end{enumerate}

\noindent\textbf{Advantage-related filtering score.}
Throughout the paper, we use $\hat{A}$ to denote a prompt-level advantage-related filtering score rather than the per-response GRPO advantage $A_{i,t}$. The estimator is supervised with the empirical Avg@$G$ score
\begin{equation}
A(q)=\frac{1}{G}\sum_{i=1}^{G}R_i,
\end{equation}
which serves as a proxy for identifying prompts that are likely to yield zero- or low-variance GRPO advantages. Prompts with predicted score $\sigma(\hat{A}(q))$ close to $0$ or $1$ are likely to produce all-incorrect or all-correct response groups and are therefore filtered before rollout.

\subsubsection{Training Objective}
To ensure that the estimator learns a robust and discriminative estimation of the advantage-related filtering scores, we employ a joint loss function that combines three distinct training objectives:
\begin{equation*}
    \mathcal{L} = \mathcal{L}_{\text{DE}} + w_{\text{distill}} \cdot \mathcal{L}_{\text{distill}} + w_{\text{rank}} \cdot \mathcal{L}_{\text{rank}}
\end{equation*}
where $w_{\text{distill}}$ and $w_{\text{rank}}$ are hyperparameters that balance the contribution of each component.

\paragraph{Advantage Estimation Loss}
$\mathcal{L}_{\text{DE}}$ trains the model to predict the ground-truth score $A$. We adopt a Binary Cross-Entropy (BCE) loss:
\begin{align*}
    \mathcal{L}_{\text{DE}} = -\big[ & A \log(\sigma(\hat{A})) \\
    & + (1 - A) \log(1 - \sigma(\hat{A})) \big]
\end{align*}

In the following loss, $A$ denotes the ground-truth Avg@$G$ target used as a proxy for advantage-based filtering, and $\hat{A}$ denotes its prediction.

We found BCE to empirically outperform the conventional Mean Squared Error (MSE) loss, $\mathcal{L}_{\text{MSE}} = \frac{1}{2}(A-\sigma(\hat{A}))^2$, and attribute this to their gradient dynamics:
\begin{align*}
    \frac{\partial \mathcal{L}_{\text{MSE}}}{\partial \hat{A}} &= (\sigma(\hat{A})-A)\,\sigma(\hat{A})(1-\sigma(\hat{A})), \\
    \frac{\partial \mathcal{L}_{\text{BCE}}}{\partial \hat{A}} &= \sigma(\hat{A})-A .
\end{align*}
The MSE gradient carries the factor $\sigma(\hat{A})(1-\sigma(\hat{A}))$. 
This is precisely the extreme range we must resolve to identify trivial and intractable prompts. The BCE gradient instead depends solely on the prediction error, supplying a consistent corrective signal at any magnitude.

\paragraph{Distillation Loss}
To further align the estimator with the actor's current capabilities, we introduce a distillation loss. This objective tasks the estimator with predicting the actor model's perplexity $P$ on the given question. By distilling this knowledge from the actor, the estimator gains a more nuanced, actor-centric understanding of difficulty on the given problem. Empirically, we found this auxiliary task improves the model's utility for zero-variance problem filtering. We also use BCE loss for this objective after normalizing the PPL values:
\begin{align*}
    \mathcal{L}_{\text{distill}} = -\big[ & P \log(\sigma(\hat{P})) \\
    & + (1 - P) \log(1 - \sigma(\hat{P})) \big]
\end{align*}

\paragraph{Ranking Loss}
Regression models commonly collapse toward the dataset mean, which here would compress most predicted scores into a narrow interval and make the easiest and hardest prompts indistinguishable. Since harder prompts should have lower predicted Avg@$G$ scores, we define a difficulty score
\begin{equation}
\hat{d}=1-\sigma(\hat{A}),
\end{equation}
and apply a pairwise ranking loss to $\hat{d}$:
\begin{equation}
L_{rank}= \frac{1}{|Q|}\sum_{(i,j)\in Q}\max(0, m-(\hat{d}_i-\hat{d}_j))
\end{equation}
where $Q$ is a set of training pairs $(i,j)$ with $i$ known to be harder than $j$, and $m$ is a margin. The loss penalizes the model whenever the harder prompt's difficulty score fails to exceed the easier one's by at least $m$, blocking trivial mean prediction.

\subsection{``Cold-Start'' Problem of the Difficulty Estimator}\label{sec:estimator_warmup}
The estimator starts from pre-trained BERT weights, but filtering with it immediately would be counterproductive: before task-specific calibration its predictions are noisy enough to discard high-value samples. We adopt a two-stage schedule:
\begin{itemize}
    \item \textbf{Estimator Warm-up Phase}: During the first \(100\) steps, filtering is disabled. The actor follows standard GRPO, while the estimator is trained on the resulting rollouts using the objective in Section~\ref{sec:estimator}.
    \item \textbf{Active Filtering Phase}: Once the warm-up phase is complete and the estimator has achieved a baseline level of convergence, the filtering mechanism is activated. The estimator subsequently generates advantage-related filtering scores to dynamically filter out prompts that are likely to produce zero- or low-variance rewards.
\end{itemize}

\begin{figure}[t]
    \centering
    \setlength{\belowcaptionskip}{-0.3cm}
    \includegraphics[width=0.95\columnwidth]{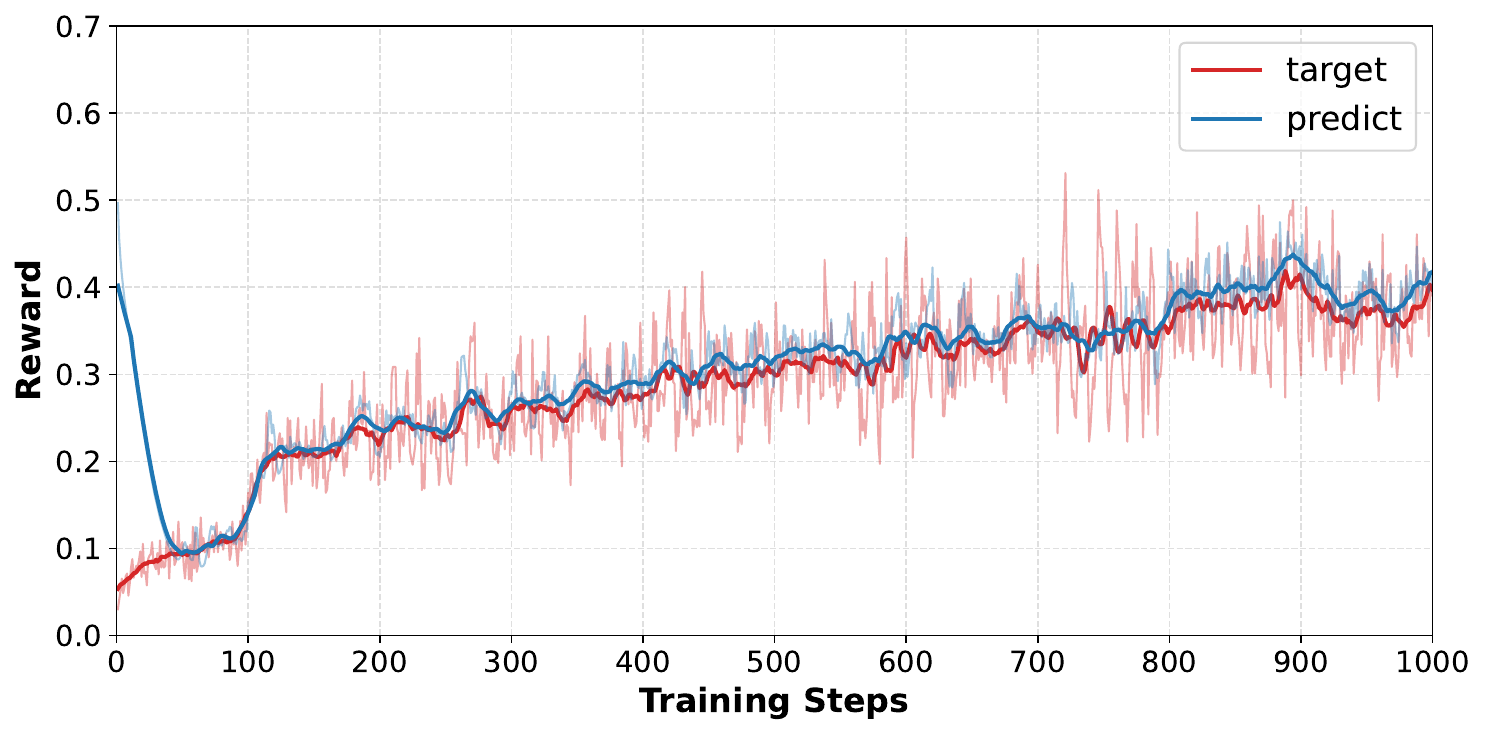}
    \caption{The comparison between the predicted rewards from the estimator and the ground-truth target rewards derived from the actor model. The estimator effectively converges, demonstrating a high degree of fidelity in tracking the target reward trajectory throughout the training process.}
    \label{fig:estimated_vs_gt_adv}
\end{figure}

Figure~\ref{fig:estimated_vs_gt_adv} illustrates the temporal alignment between the rewards predicted by the Difficulty Estimator and the ground-truth rewards generated by the actor model across the training trajectory. Following a brief 100-step warm-up phase, the estimator exhibits stable convergence, demonstrating a high degree of fidelity in tracking the target reward distributions. This accurate approximation is foundational to the efficacy of the subsequent advantage-based filtering mechanism, as it ensures that the model prioritizes informative samples based on a reliable proxy of task difficulty.

\begin{table*}[t]
\setlength{\tabcolsep}{3pt}
\footnotesize
\centering
\setlength{\belowcaptionskip}{-0.3cm}
    \begin{tabular}{c|l|ccccc|c|c}
    \toprule
    \textbf{Dataset} & \textbf{Method} & \textbf{GSM8K} & \textbf{MATH} & \textbf{AMC23} & \bf Olympiad & \bf Minerva & \textbf{Avg.} & \textbf{GPU Hours $\downarrow$} \\
    \midrule
    \midrule
    \multirow{12}{*}{\rotatebox[origin=c]{90}{\bf \textsc{DAPO-MATH-17K}}} & \multicolumn{8}{c}{\bf\emph{Qwen2.5-1.5B-Instruct}} \\
    \cmidrule{2-9}
    & \grpo & 75.6 & 48.1 & 38.4 & 15.8 & 11.4 & 37.9 & \bf 528 (1.0$\times$) \\
    & \dapo & \bf 78.5 & 50.1 & 39.3 & \bf 17.8 & 13.1 & \underline{39.8} & 905 (1.7$\times$) \\
    & \polaris & 77.1 & 47.3 & 40.8 & 16.4 & 11.8 & 38.7 & 584 (1.1$\times$) \\
    \cmidrule{2-9}
    & \bf \depo & 77.0 & 48.9 & \bf 42.3 & 16.7 & 12.2 & 39.4 & \underline{530 (1.0$\times$)} \\
    & \quad -- ranking loss & 76.6 & 48.0 & 40.9 & 16.3 & 12.1 & 38.8 & \underline{530 (1.0$\times$)} \\
    & \quad \quad -- distill loss & 75.2 & 48.0 & 39.0 & 15.9 & 12.0 & 38.0 & \underline{530 (1.0$\times$)} \\
    & \quad + \dapo w/o Dyn. Samp. & 78.3 & \bf 50.6 & 41.7 & 17.5 & \bf 13.3 & \bf 40.3 & \underline{530 (1.0$\times$)} \\
    \cmidrule{2-9}
    \cmidrule{2-9}
    & \multicolumn{8}{c}{\bf\emph{Qwen2.5-7B-Instruct}} \\
    \cmidrule{2-9}
    & \grpo & 91.9 & \bf 64.1 & 63.4 & 27.9 & 25.0 & 54.5 & \bf 776 (1.0$\times$) \\
    & \depo & \bf 92.3 & \bf 63.9 & \bf 63.5 & \bf 28.7 & \bf 25.5 & \bf 54.8 & 782 (1.0$\times$) \\
    \midrule
    \multirow{3}{*}{\rotatebox[origin=c]{90}{\bf \textsc{OR1}}} & \multicolumn{8}{c}{\bf\emph{Qwen2.5-7B-Instruct}} \\
    \cmidrule{2-9}
    & \grpo & \bf 92.0 & 63.3 & 48.9 & 26.4 & 26.2 & 51.4 & \bf 776 (1.0$\times$) \\
    & \depo & 91.8 & \bf 64.0 & \bf 51.0 & \bf 27.6 & \bf 26.6 & \bf 52.2 & 782 (1.0$\times$) \\
    \midrule
    \multirow{3}{*}{\rotatebox[origin=c]{90}{\bf \textsc{NT}}} & \multicolumn{8}{c}{\bf\emph{Qwen2.5-7B-Instruct}} \\
    \cmidrule{2-9}
    & \grpo & 90.1 & 62.7 & 48.9 & 25.3 & 23.8 & 50.1 & \bf 776 (1.0$\times$) \\
    & \depo & \bf 90.8 & \bf 63.2 & \bf 53.2 & \bf 25.6 & \bf 25.0 & \bf 51.6 & 782 (1.0$\times$) \\
    \bottomrule
    \end{tabular}
    \caption{Performance comparison (Avg@32) across five math reasoning benchmarks. \depo achieves performance comparable to \dapo while significantly reducing training overheads.}
    \label{tab:main_results}
\end{table*}

\section{Experiments}
\subsection{Experimental Settings}
\paragraph{Models \& Datasets.}
We run our experiments on Qwen2.5-1.5B-Instruct and Qwen2.5-7B-Instruct~\citep{qwen2025qwen25technicalreport}. Our method and all baselines are trained on three datasets: DAPO-MATH-17K~\citep{yu2025dapoopensourcellmreinforcement}, OPEN-R1 (OR1)~\citep{openr1}, and Nemotron-Math~\citep{wang2025nemotroncascadescalingcascadedreinforcement}.

\paragraph{Training \& Evaluation Details.}
Our method and all baselines are implemented using the Verl framework~\citep{Sheng_2025}, with vLLM~\citep{kwon2023efficient} as the backend for rollouts. We use 16xH100 for Qwen2.5-1.5B-Instruct training and 32xH100 for Qwen2.5-7B-Instruct. To ensure a rigorous comparison under constrained computational resources, all models were trained for 1,000 steps with a global batch size of 128 and a learning rate of $1e-6$. We generate $G=8$ rollouts for each prompt during training. Following the warmup strategy described in Section~\ref{sec:estimator_warmup}, the first 100 steps were dedicated to training the difficulty estimator in isolation, and the filtering mechanism remains inactive during this phase. We empirically set $W_{\text{distill}}$ and $W_{\text{ranking}}$ to 0.5 and 3, respectively.We retain prompts whose predicted success probabilities fall within $[0.2, 0.8]$, using a heuristic band to prioritize moderate-difficulty prompts under noisy estimates. This rule does not imply that every discarded prompt would produce a zero-variance response group.


For evaluation datasets, we include five mathematical reasoning benchmarks: GSM8K~\citep{cobbe2021trainingverifierssolvemath},  MATH~\citep{hendrycks2021measuringmathematicalproblemsolving}, AMC23~\citep{amc23}, Minerva
Math~\citep{lewkowycz2022solvingquantitativereasoningproblems}, and Olympiad Bench~\citep{he-etal-2024-olympiadbench}. During inference, we use vLLM with a temperature of $T=1$ and $top\_p=0.95$ for nucleus sampling. For each test instance, we generate 32 responses and report the $\text{Avg}@32$ as the primary metric for comparison.

\subsection{Main Results}
\paragraph{\depo achieves superior performance over \grpo while maintaining high computational efficiency.}
As shown in Table~\ref{tab:main_results}, \depo improves on the \grpo baseline by 1.5\% on average, remaining competitive with far more resource-intensive baselines while matching \grpo's efficiency --- a strictly better accuracy/throughput Pareto frontier than \dapo, which pays substantial extra compute for its gains.

\paragraph{\depo is complementary with existing frameworks.}
Being plug-and-play, \depo composes with existing RL methods. Substituting our filter for \dapo's dynamic sampling (\ie \dapo without dynamic sampling, plus \depo) yields a further 0.9\% average gain, confirming that \depo is complementary to state-of-the-art frameworks rather than competing with them.

\paragraph{Synergy of Ranking and Distillation loss.}
Removing either the ranking or the distillation loss consistently degrades downstream performance, confirming that both are needed to calibrate the Difficulty Estimator. Without them the estimator fails to discriminate low-utility prompts, readmitting the stochastic noise and gradient sparsity that stall optimization.

\begin{figure}[t]
    \centering
    \setlength{\belowcaptionskip}{-0.3cm}
    \includegraphics[width=0.95\columnwidth]{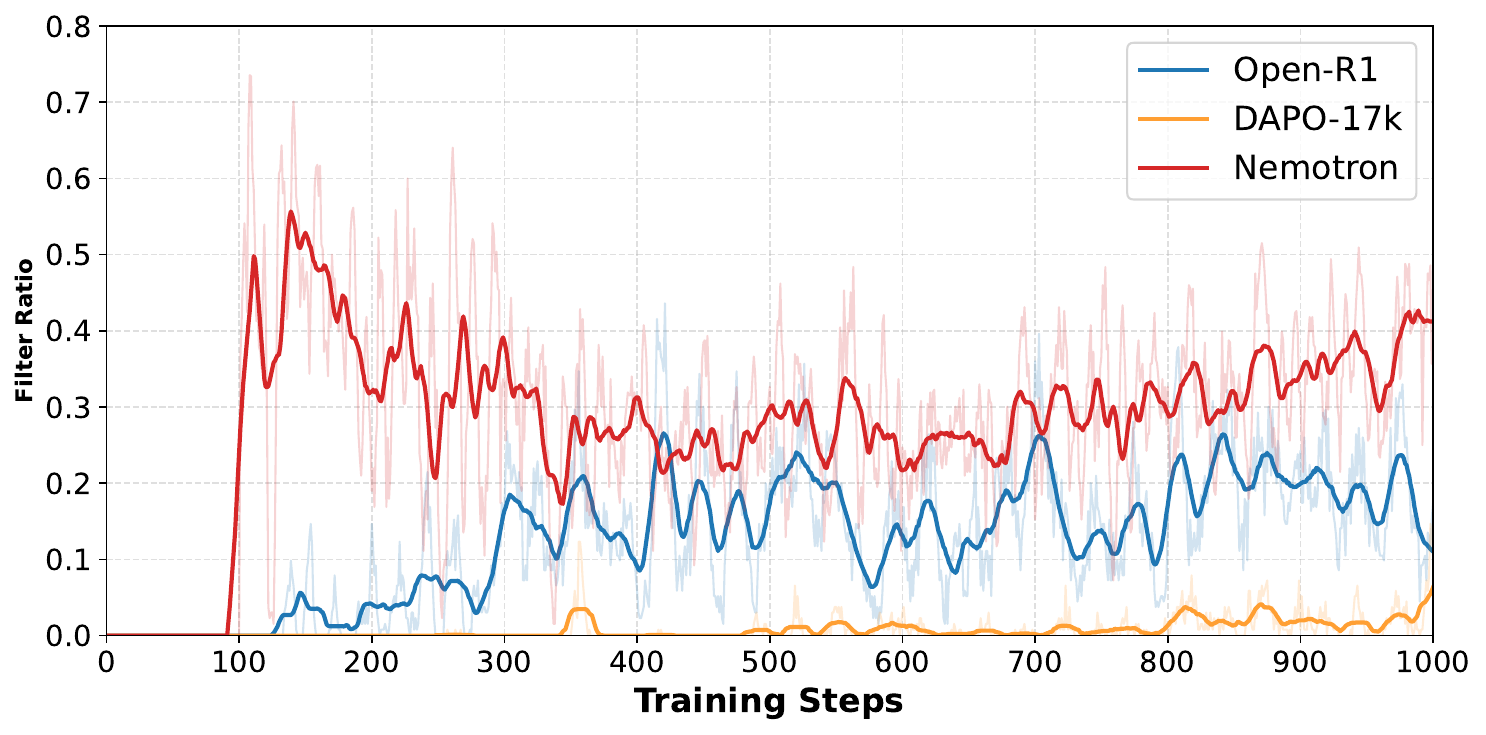}
    \caption{Dynamics of filtering ratios when training Qwen2.5-7B-Instruct on datasets of varying difficulty.}
    \label{filter_ratio_different_datasets}
\end{figure}

\paragraph{\depo is sensitive to model capability and training dataset difficulty.}
\depo's efficacy depends on both model capability and the difficulty distribution of the training data. On Qwen2.5-7B-Instruct with DAPO-MATH-17K the gains are marginal, whereas Open-R1 (easier) and Nemotron-Math (harder) yield markedly larger improvements. The gain is therefore largest when the data holds many low-utility samples (\eg trivial or intractable for the current policy) and smallest when most samples sit near the model's capability and already provide useful gradients. Figure~\ref{filter_ratio_different_datasets} confirms this: filtering ratios for Open-R1 and Nemotron-Math are well above that of the more moderately distributed DAPO-MATH-17K.

\subsection{Analysis}

\begin{table}[t]
\setlength{\tabcolsep}{3pt}
\footnotesize
\centering
\setlength{\belowcaptionskip}{-0.3cm}
\begin{tabular}{l|ccccc}
    \toprule
    \textbf{Method} & \textbf{Sample} & \textbf{Rollout} & \shortstack{\textbf{Adv.}\\\textbf{Comput.}} & \textbf{Reward} & \textbf{Total $\downarrow$} \\ \midrule
    \grpo & 74.63 & 103.75 & 0.021 & 0.406 & \bf 121.85 \\
    \dapo & 76.98 & 192.16 & 0.067 & 2.064 & 211.69 \\
    \depo & 74.99 & 103.11 & 0.186 & 0.403 & \underline{125.65} \\ \bottomrule
\end{tabular}
\caption{Runtime breakdown and efficiency comparison per training step (seconds). \depo achieves a $1.86 \times$ speedup in rollout efficiency compared to \dapo, maintaining a total step latency comparable to the \grpo baseline.}
\label{tab:runtime}
\end{table}

\paragraph{\depo achieves efficiency comparable to \grpo and 1.86 $\times$ speedup over \dapo.}
Table~\ref{tab:runtime} breaks down per-step training time. Sampling latency is essentially uniform across all three methods, but rollout duration varies sharply: \dapo's over-sampling makes its rollouts nearly twice as costly. \depo's per-step latency stays comparable to \grpo at roughly 41.4\% of \dapo's overhead.

\begin{figure}[t]
    \centering
    \setlength{\belowcaptionskip}{-0.3cm}
    \includegraphics[width=0.9\columnwidth]{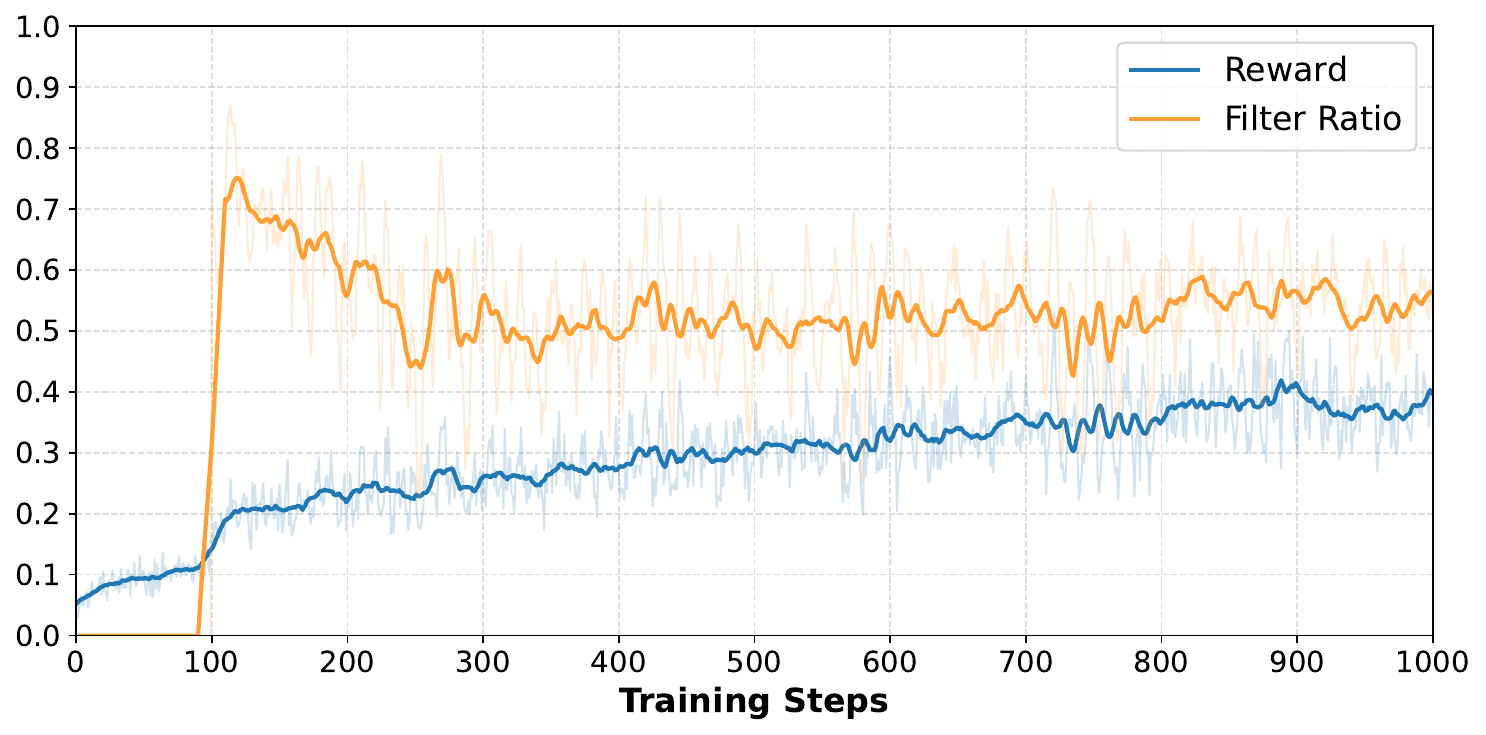}
    \caption{Training dynamics of mean rewards and prompt filtering ratios across the training trajectory. We observe that half of the non-informative prompts are filtered out on average, significantly improving the training efficiency.}
    \label{fig:reward_filter_ratio}
\end{figure}

\paragraph{Difficulty Estimator results in high filtering ratio.}
We define the filtering ratio as the fraction of sampled candidate prompts not selected for policy optimization. Figure~\ref{fig:reward_filter_ratio} shows that once filtering activates after warm-up, mean reward rises substantially and consistently while the filtering ratio settles near 41\% for the rest of training. The mechanism thus prunes roughly half of all prompts as zero-variance, eliminating the full-rollout cost they would otherwise incur.

\begin{figure}[t]
    \centering
    \setlength{\belowcaptionskip}{-0.3cm}
    \includegraphics[width=0.9\columnwidth]{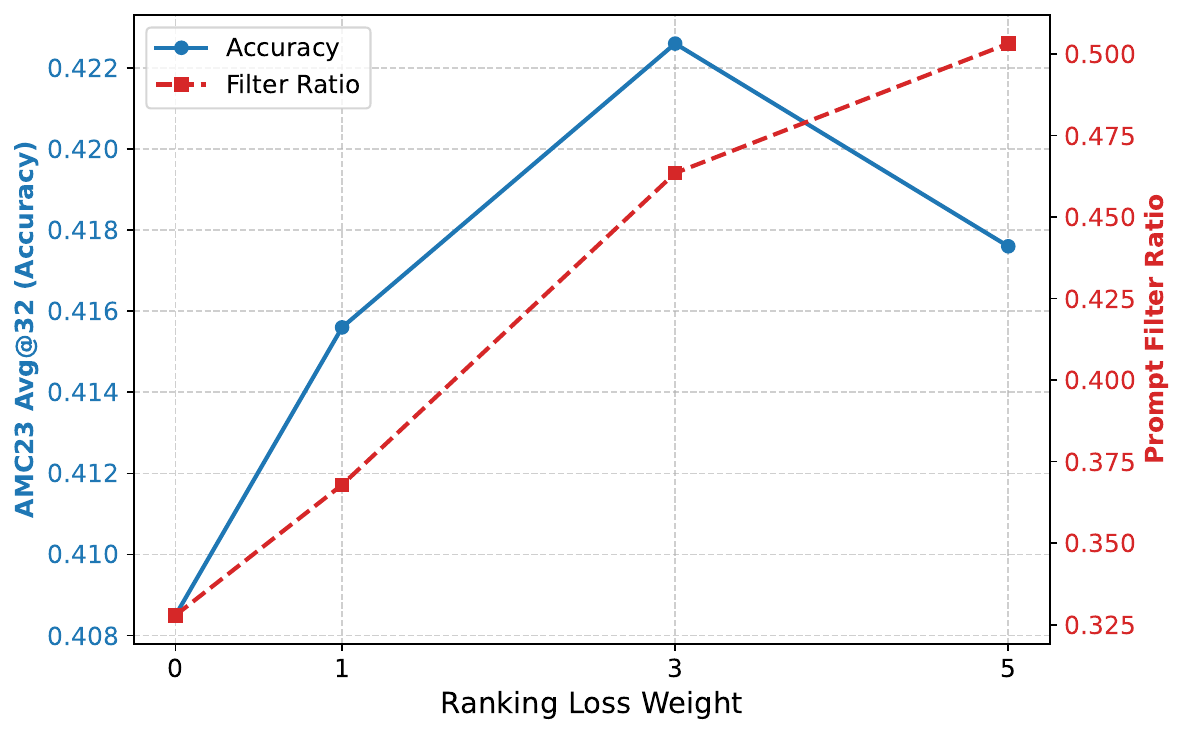}
    \caption{Impact of ranking loss weight on downstream accuracy and prompt filtering efficiency. A dual effect of increasing ranking loss weight on model performance and the prompt filtering ratio is observed, where the accuracy reaches a peak while the filter ratio increases monotonically with higher weights.}
    \label{fig:ranking_loss_ablation}
\end{figure}

\paragraph{Ranking loss leads to more aggressive prompt filtering.}
We ablate the weight of $\mathcal{L}_{rank}$ to probe \depo's sensitivity. Figure~\ref{fig:ranking_loss_ablation} reveals a clear trade-off: accuracy improves as the weight increases and filtering grows more selective, but performance degrades after pasting an optimal threshold, around which roughly 41\% of prompts are discarded. Excessive weights therefore induce over-filtering, pruning marginal but still beneficial samples and reducing the diversity of training signals.

\begin{figure}[t]
    \centering
    \setlength{\belowcaptionskip}{-0.3cm}
    \includegraphics[width=0.9\columnwidth]{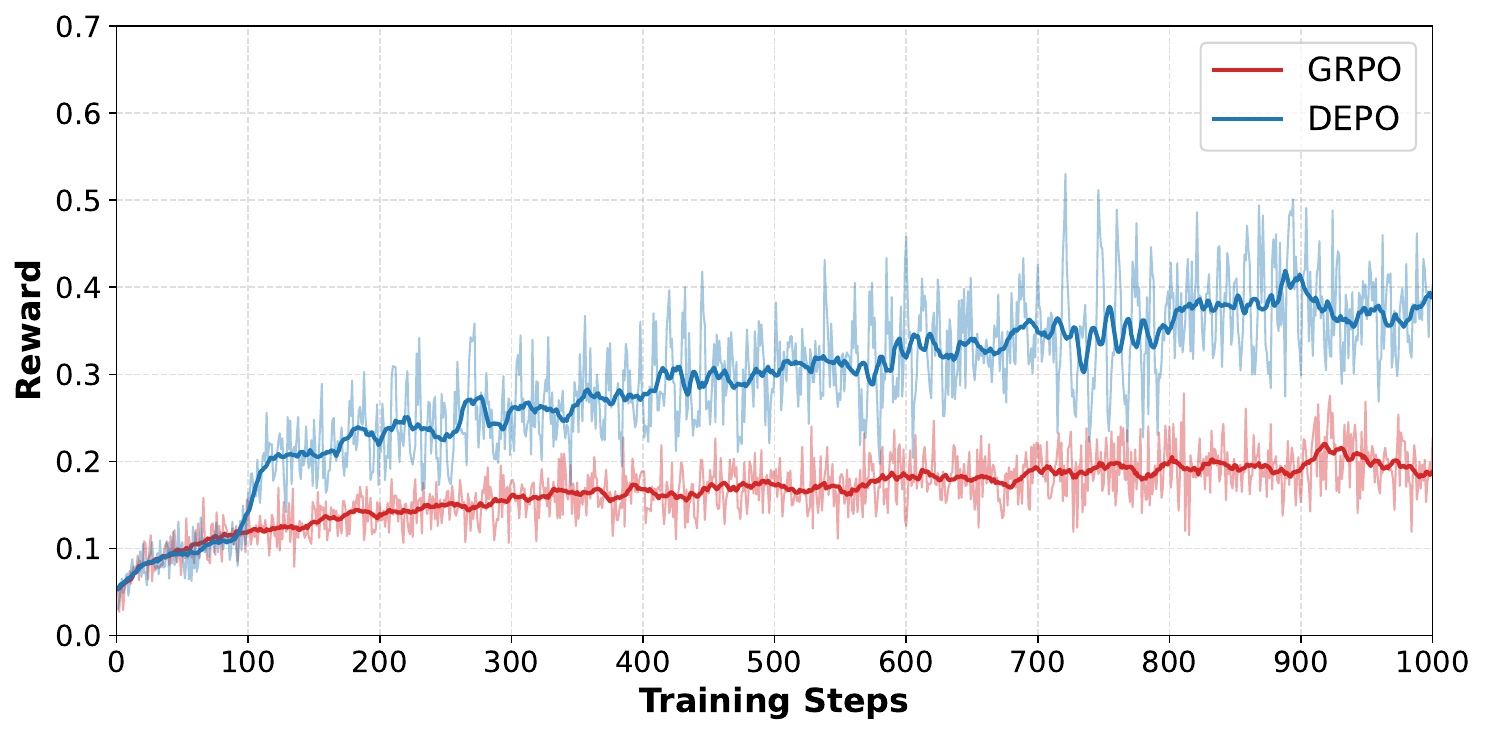}
    \caption{The comparison of rewards attained by the actor model by employing \grpo and \depo. We observe the model trained using \depo consistently obtains higher rewards than that of using \grpo.}
    \label{fig:grpo_vs_depo_rewards}
\end{figure}

\paragraph{Rewards observed higher in \depo than \grpo.}
Comparing the rewards received by the actor under each method (Figure~\ref{fig:grpo_vs_depo_rewards}), we find that \depo rises sharply once filtering activates and that its margin over \grpo widens monotonically across the remaining trajectory. This indicates that the filtering mechanism successfully excludes intractable prompts (\ie reward $=0$), enriching each batch with informative signals for policy optimization.

\begin{table*}[t]
\setlength{\tabcolsep}{6pt}
\footnotesize
\centering
\setlength{\belowcaptionskip}{-0.3cm}
\begin{tabular}{l|c|ccccc|c|c}
    \toprule
    \bf Model & $\tau$ & \bf GSM8K & \bf MATH & \bf AMC23 & \bf Olympiad & \bf Minerva & \bf Avg. & $\Delta$ \\
    \midrule
    1.5B & - & 77.0 & 48.9 & 42.3	& 16.7 & 12.2 & 39.4 & - \\
    7B & - & \bf 92.3 & \bf 63.9 & \underline{63.5} & \bf 28.7 & \bf 25.5 & \bf 54.8 & - \\
    \midrule
    \multirow{2}{*}{1.5B + 7B} & \multirow{2}{*}{0.75} & \underline{87.1} & 62.4 & \bf 65.0 & \underline{28.0} & \underline{25.1} & \underline{53.5} & +14.1/-1.3 \\
    & & 557/762 & 129/371 & 2/38 & 27/647 & 35/237 & 750/2805 & 26.7\% \\
    \midrule
    \multirow{2}{*}{1.5B + 7B} & \multirow{2}{*}{0.5} & 82.0 & 57.5 & 61.3 & 25.3 & 20.6 & 49.4 & +10.0/-5.4 \\
    & & 997/322 & 303/197 & 7/33 & 117/557 & 125/147 & 1549/2805 & 55.2\% \\
    \midrule
    \multirow{2}{*}{1.5B + 7B} & \multirow{2}{*}{0.3} & 79.6 & 53.5 & 59.1 & 24.4 & 17.9 & 46.9 & +7.5/-7.9 \\
    & & 1162/157 & 370/130 & 12/28 & 175/499 & 197/93 & 1916/2805 & 68.3\% \\
    \bottomrule
\end{tabular}
\caption{The results when employing the difficulty estimator to dynamically route incoming queries to different models. $\Delta$ indicates the performance difference when comparing to solely using 1.5B/7B models. The number of queries processed by 1.5B and 7B models are presented respectively.}
\label{tab:online_router_results}
\end{table*}

\section{Difficulty Estimator as Model Router}
We implement an online routing mechanism to assess how our difficulty estimator can facilitate the cooperation of heterogeneous models with varying capabilities. By utilizing the estimator as a router, we implement a routing pipeline where the complexity of the task dictates the model size. In this setup, the routing logic is governed by a predefined confidence threshold $\tau$. For any incoming query, the difficulty score is first computed; if the model's confidence is deemed insufficient (\ie $\text{score} < \tau$), the system assumes the query exceeds the reliable capacity of the smaller, more efficient model. Consequently, such hard queries are routed to a high-capacity model. This cascaded approach allows for an optimized trade-off between inference latency and final performance.

As shown in Table~\ref{tab:online_router_results}, the proposed routing mechanism achieves performance competitive with the 7B model while offloading 27\% of queries to the 1.5B model with negligible degradation. By lowering the threshold $\tau$, thereby increasing the proportion of queries processed by the smaller model, the system can handle up to 68\% of the workload via the 1.5B model while still yielding an 8\% improvement in averaged accuracy over the 1.5B model. These findings underscore the efficacy of repurposing the difficulty estimator as an online, zero-shot router to effectively balance downstream accuracy and computational efficiency.

\section{Related Work}
\subsection{RL for LLM Reasoning}
RLHF~\citep{ouyang2022traininglanguagemodelsfollow} and RLVR~\citep{lambert2025tulu} were initially dominated by \ppo~\citep{schulman2017proximalpolicyoptimizationalgorithms}, which requires a four-model architecture (actor, critic, reward model, and reference model) and therefore incurs substantial computational overhead and training instability. \grpo~\citep{shao2024deepseekmathpushinglimitsmathematical} removes the standalone critic by estimating advantages from a group of sampled outputs; with rule-based rewards for verifiable tasks such as mathematics, this reduces to a two-model framework (\ie actor and reference), lowering the resource barrier considerably.

Building on \grpo, recent work refines the quality of training signals. \dapo~\citep{yu2025dapoopensourcellmreinforcement} introduces dynamic sampling to counter vanishing advantages; PRIME~\citep{cui2025processreinforcementimplicitrewards} derives Implicit Process Rewards from labels for denser gradients; FAPO~\citep{ding2025fapoflawedawarepolicyoptimization} separates \emph{fully correct} from \emph{flawed but correct} trajectories for more precise supervision; and Dr.GRPO~\citep{liu2025understandingr1zeroliketrainingcritical} removes length and standard-deviation normalization biases. These refinements improve how the model learns from rollouts but leave rollout costs untouched. \depo targets data-level efficiency by filtering instances before rollout, making it orthogonal to and combinable with all of the above.

\subsection{Data Curation for LLM Reasoning}
A complementary line of work curates training data, either by static pre-filtering or dynamic calibration. Static methods score sample utility before training: LoBaSS~\citep{zhou2024davirdataselectionimplicit} uses the perplexity delta across a training stage to identify \emph{learnable} samples, while ScalingFilter~\citep{li-etal-2024-scalingfilter} uses the perplexity gap between a small proxy and a larger target model. Polaris~\citep{Polaris2025} shows that reasoning difficulty follows a \emph{J-shaped distribution} and that aligning data to a target difficulty profile accelerates convergence. Being decoupled from the actor's evolving state, however, such static estimates grow stale as the policy shifts.

Dynamic strategies adapt prompt selection during training. \citet{sun2025improvingdataefficiencyllm} uses a predictor trained on separately collected difficulty supervision and performs online rollouts on a reference set to estimate the difficulty of other candidates. \depo instead trains a text-based BERT estimator online using rewards and actor-derived perplexity targets from the ongoing policy-training loop. It requires neither separately collected difficulty-supervision data nor additional reference-set rollouts for candidate scoring.


\section{Conclusion}
In this paper, we present \depo to address a critical computational bottleneck in training LRMs: the high cost of rollouts for low-utility training samples. By integrating an online Difficulty Estimator, our framework proactively filters trivial or intractable prompts before the expensive rollout phase. Empirical results demonstrate that \depo achieves a 50\% reduction in total computational overhead compared to SOTA frameworks like \dapo, all while outperforming the standard \grpo baseline. Its plug-and-play nature and fundamental orthogonality to other algorithmic advancements make it a sustainable and scalable solution for advancing the frontier of reasoning-heavy reinforcement learning.

\section*{Limitations}
Our experiments are limited to mathematical reasoning with verifiable rewards and to Qwen2.5-Instruct models at the 1.5B and 7B scales, trained for 1,000 steps under a constrained compute budget. Whether \depo transfers to other domains (\eg code generation or tasks without verifiable rewards), other model families, larger models, and longer training horizons remains to be verified. The benefit of \depo also depends on the difficulty distribution of the training data: when most prompts already lie near the actor's capability, as for Qwen2.5-7B-Instruct on DAPO-MATH-17K, the gains are marginal. The Difficulty Estimator sees only the question text and is supervised with the Avg@$G$ reward as a proxy for the advantage, so its errors can discard informative prompts; our ablation of the ranking-loss weight shows that overly aggressive filtering hurts accuracy.

\bibliography{custom}

\end{document}